\documentclass{IOS-Book-Article}
\usepackage{CJKutf8}
\usepackage{mathptmx}
\usepackage{amsmath}
\usepackage{graphicx}
\usepackage{soul}\setuldepth{article}
\def\hb{\hbox to 11.5 cm{}}

\begin{document}

\begin{CJK}{UTF8}{bsmi} 

\pagestyle{headings}
\def\thepage{}
\begin{frontmatter}              

\title{Similar Phrases for Cause of Actions of Civil Cases}

\markboth{}{September 2024\hb}

\author{\fnms{Ho-Chien} \snm{Huang}}$\dagger$
\qquad \qquad \qquad \qquad \qquad
\author{\fnms{Chao-Lin} \snm{Liu}}$\ddagger$
\address{$\dagger$College of Law, National Chengchi University, Taiwan}
\address{$\dagger \ddagger$Department of Computer Science, National Chengchi University, Taiwan}
\texttt{\{$\dagger$110601014, $\ddagger$chaolin\}@g.nccu.edu.tw}

\begin{abstract}
In the Taiwanese judicial system, Cause of Actions (COAs) are essential for identifying relevant legal judgments. However, the lack of standardized COA labeling creates challenges in filtering cases using basic methods. This research addresses this issue by leveraging embedding and clustering techniques to analyze the similarity between COAs based on cited legal articles. The study implements various similarity measures, including Dice coefficient and Pearson's correlation coefficient. An ensemble model combines rankings, and social network analysis identifies clusters of related COAs. This approach enhances legal analysis by revealing inconspicuous connections between COAs, offering potential applications in legal research beyond civil law.
\end{abstract}

\end{frontmatter}
\markboth{September 2024\hb}{September 2024\hb}

\section{Introduction}
In the Taiwanese judicial system, case assigners allocate cases to judges after receiving declarations from plaintiffs. Each case is tagged with a \textit{Cause of Action (COA)}, which encapsulates its key issues or claims. These COAs are essential for legal professionals, such as lawyers and legal assistants, in identifying relevant judgments. Since COAs are manually labeled and carry significant legal meaning, cases sharing the same COA often exhibit similar claims, cited articles, and defense strategies.

Leveraging COAs helps filter out irrelevant judgments and resolves ambiguities between similar terms, such as 僱傭 and 雇用 (both meaning employment). For instance, Shao \emph{et al.} used the keywords ``酒" (alcohol) and ``刑法185-3" (Criminal Code Art. 185-3) to identify drunk driving cases \cite{r1}. However, a keyword-based approach can miss relevant cases that do not contain specific terms and can include irrelevant ones.

The absence of a standardized system for assigning COAs in the Taiwanese judiciary leads to synonymous COAs, making it challenging to identify similar cases using basic methods such as minimum edit distance or clustering based solely on COA labels. For example, 返還借款 (loan repayment) and 返還借貸款 (loan repayment) both refer to the same type of case, while 返還貨款 (repayment of goods) emphasizes repayment for goods. This inconsistency complicates the work of legal analysts when filtering cases or ensuring dataset comprehensiveness. For instance, Liu \emph{et al.} filtered cases using COAs containing ``扶養費" (alimony) \cite{r2}, potentially missing synonymous COAs such as 酌定扶養方法 (determined maintenance method).

With over 14,300 COA types, manually determining their similarity is impractical. Since COAs are labeled by case assigners, cases with identical COAs tend to have similar claims. As such, they are more likely to cite the same articles. In our research, we use the co-citation of articles to measure the similarity between COAs. We further employ embedding and clustering techniques to analyze plaintiffs’ claims and determine thresholds of COA similarity.

This research aims to improve the completeness and accuracy of the work of legal professionals and analysts. Additionally, it seeks to support case assigners in their decision-making processes by incorporating machine learning techniques.


\section{Problem Definition}

The goal of this study is to find out the sets of similar phrases for \textit{cause of actions} that were used in the judgment documents of the civil cases in Taiwan. Normally, we call the cause of actions as \textit{titles} (案由) in Chinese, and, as a trade-off, we will use \textbf{COAs} as a shorthand for \textit{Cause of Actions} in our presentation. 

We will measure the similarity between a given pair of COAs based on the articles that were cited in the judgment documents of the cases that belonged to the COAs being compared.

Let $m$ denote the number of different COAs in the current study. We denote the set of COAs as $\mathbf{T} = \{T_1, T_2, \ldots,T_i
, \ldots, T_m\}$. Let $n$ denote the number of cases that we sampled for each of the COAs. Then, for a $T_i$, we denote the cases for $T_i$ as $\mathbf{S_i} = \{S_{i1}, S_{i2}, \ldots, S_{ij}, \ldots, S_{in}\}$, where a $S_{ij}$ is the \textit{j}-th sampled case of a COA $T_i$. 

We will compare the law articles that were cited in the cases to measure the similarity between individual cases, excluding some citations. There are acts, for both civil and criminal cases, that govern how the lawsuits should proceed in general. They are \textit{Taiwan Code of Civil Procedure} (民事訴訟法) and \textit{Taiwan Code for Criminal Procedure} (刑事訴訟法).\footnote{We use English translations for acts, codes, and laws of Taiwan from this site: https://www.judicial.gov.tw/tw/lp-158-1-10-40.html. The Judicial Yuan (司法院), the highest government unit for judicial matters, maintains this site.} Citing articles in these acts may not be relevant to the main issues of the lawsuits. Hence, we ignore citations to these procedural regulations in our comparison procedure.

A civil case may cite different numbers of articles from different acts. Each of these different articles will be assigned a different internal code so that we can compare the articles cited for a given pair of cases precisely. For instance, a case A may site two articles, 229 and 233, in the \textit{Civil Code} (民法) and the article 125 in the \textit{Banking Act} (銀行法). Another case B might site article 229 in the Civil Code and article 125 in the Banking Act. Then our comparison procedure would identify that both cases A and B cite two articles. 

After we sample $n$ cases for each of $m$ COAs, we would have $mn$ cases. We can use a computer program to find out the articles that were cited in these $mn$ cases, excluding the procedure regulations. Let us assume that there are $\alpha$ different articles cited by these $mn$ cases. We denote the set of cited articles as $\mathbf{A} = \{A_1, A_2, \ldots, A_k, \ldots, A_\alpha\}$.

Therefore, we may convert a sampled case to a vector of $\alpha$ 1s and 0s, where $A_k$ is 1 or 0 would indicate that a case sites article $A_k$ or not, respectively. For instance, we may say that $ v(S_{ij})= (1, 0, 1, ⋯ , 0)$ if $S_{ij}$ sites articles $A_1$ and $A_3$. 

To measure the similarity between two COAs, $T_x, T_y \in \mathbf{T}$, we will define and choose some possible similarity measures for $sim(T_x, T_y)$. The similarity between $T_x$ and $T_y$ will be determined by the similarity between their sampled cases, $sim(\mathbf{S_x},\mathbf{S_y})$, and we will provide more details about $sim(\mathbf{S_x},\mathbf{S_y})$ shortly.

\begin{equation}
sim(T_x, T_y) \equiv sim(\mathbf{S_x},\mathbf{S_y})
\end{equation}

\section{Data Source}

We have described the major steps for data preparation in previous papers in JURIX 2022 \cite{r3} and JURISIN 2023 \cite{r2}, so we will not repeat all of the details to avoid the concerns of self-plagiarism.

We obtained the judgment documents from an open repository that is maintained by the Judicial Yuan, which is the highest government unit that governs judicial matters in Taiwan. We refer to this website as TWJY, which stands for the Taiwan Judicial Yuan.\footnote{TWJY: https://opendata.judicial.gov.tw/} The Judicial
Yuan publishes the judgment documents of several types of courts, including the local courts, the high courts, the supreme courts, and some special courts, three months after their judgment dates whenever possible. Some judgment documents may not be published
for legal reasons, e.g., to protect the minors or litigants. In addition, the contents of the documents were also anonymized for privacy reasons. The website offers documents for judicial decisions of as early as January 1996. As of April 2024, there are about 19.3 million documents available on the website. 

\section{Measuring the Similarity between COAs}

\subsection{Methodology}

Based on our explanation and definitions in Section 2, we need to define $sim(\mathbf{S_x}, \mathbf{S_y})$ to calculate the similarity between $T_x$ and $T_y$. Recall that $\mathbf{S_x}$ and $\mathbf{S_y}$ are the sampled cases of between $T_x$ and $T_y$, respectively. There are several conceivable ways to define $sim(\mathbf{S_x}, \mathbf{S_y})$.

The first one is to define $sim(\mathbf{S_x}, \mathbf{S_y})$ as the average similarity between each pair of sampled cases in $\mathbf{S_x}$ and $\mathbf{S_y}$. Recall that $v(S_{xs})$ and $v(S_{yt})$ record the articles that were cites in $S_{xs}$ and $S_{yt}$, respectively. Hence, one common choice to define $sim(v(S_{xs}), v(S_{yt}))$ is to define it as the Dice coefficient 
 \cite{r4} between the sets of articles of that were cited by $S_{xs}$ and $S_{yt}$.

\begin{equation}
sim(\mathbf{S_x}, \mathbf{S_y}) 
= \frac{1}{n^2}\sum^{n}_{s=1}\sum^{n}_{t=1}sim(S_{xs}, S_{yt}) 
= \frac{1}{n^2}\sum^{n}_{s=1}\sum^{n}_{t=1}sim(v(S_{xs}), v(S_{yt}))
\end{equation}

\begin{equation}
sim(v(S_{xs}), v(S_{yt})) = DICE(S_{xs},S_{yt})
\end{equation}

It is intriguing to compare $\mathbf{S_x}$ and $\mathbf{S_y}$ from a wholistic standpoint, rather than directly comparing their individual cases. This can be achieved by adding the vectors of the sampled cases of a COA to represent a COA, as shown in Eq. (4).

\begin{equation}
u(\mathbf{S_x}) = \sum^{n}_{s=1} v(S_{xs}) \quad \text{and} \quad u(\mathbf{S_y}) = \sum^{n}_{t=1} v(S_{yt})
\end{equation}

Given these wholistic representations of the COAs $\mathbf{S_x}$ and $\mathbf{S_y}$, we can compute the Pearson’s correlation coefficient \cite{r5} between the $u(\mathbf{S_x})$ and $u(\mathbf{S_y})$ as the $sim(\mathbf{S_x}, \mathbf{S_y})$.
This is shown in Eq. (5).

\begin{equation}
sim(\mathbf{S_x}, \mathbf{S_y}) = PCC(u(\mathbf{S_x}),u(\mathbf{S_y}))
\end{equation}

Alternatively, we can reset any numbers that are larger than 1 in $u(\mathbf{S_x})$ and $u(\mathbf{S_y})$ as 1, and denote these new degraded vectors as $u^\prime(\mathbf{S_x})$ and $u^\prime(\mathbf{S_y})$, respectively. Hence, the
third possible definition of $sim(\mathbf{S_x}, \mathbf{S_y})$, shown in Eq. (6), provides a wholistic perspective DICE coefficient for the similarity between $\mathbf{S_x}$ and $\mathbf{S_y}$. 

\begin{equation}
sim(\mathbf{S_x}, \mathbf{S_y}) = DICE(u^\prime(\mathbf{S_x}),u^\prime(\mathbf{S_y}))
\end{equation}

\subsection{Result of Implementing Similarity Measurement}

We sampled 200 cases for 179 COAs from the TWYJ database. In total, we sampled 35,800 cases. We also extracted the cited articles in these cases. After we excluded those procedural regulations as we noted in Section 2, these sampled cases cited 5956 different articles. Ignoring the procedural regulations did not affect the
distributions of the number of cited articles in cases significantly, as is shown in Figure 1.

\begin{figure}[h]
\includegraphics[width=8cm]{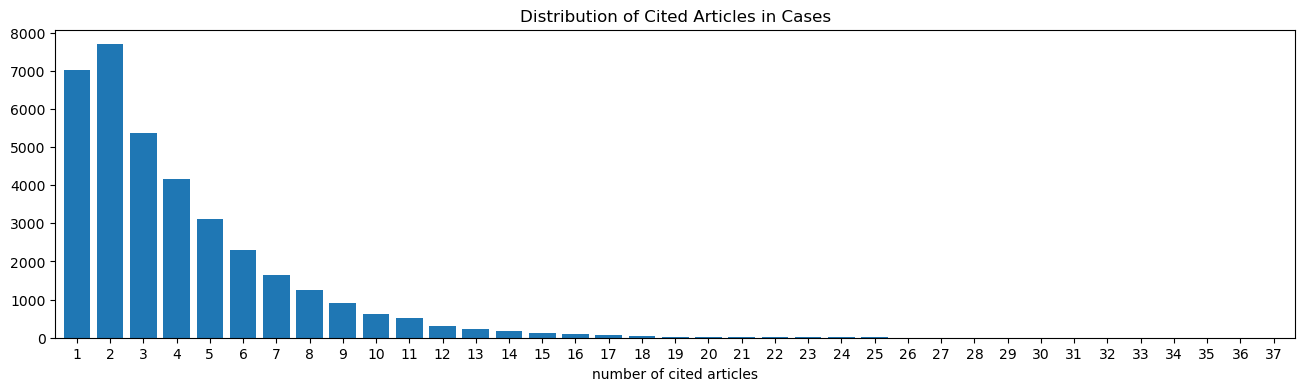}
\centering
\caption{Distributions of the number of cited articles in the sampled cases }
\end{figure}

When we computed thevsimilarity between two COAs, we did not provide the information about the COAs to our programs, our methods determined whether two COAs were similar solely based on the score functions. Hence, if our methods could identify similar COAs, we have strong evidence to support the validity of our methods.

Table~\ref{tab:rankings} lists four top-ranked pairs of COA selected by the three score functions that we discussed in Section 4.1. If you read Chinese, it should be easy to verify that these 12 pairs are similar. If you do not read Chinese, comparing the Chinese characters in the COA1 and COA2 columns may provide a reasonable impression on the similarity between the phrases in the columns.

\begin{table}[ht]
\centering
\caption{Top ranked pairs of COAs recommended}
\label{tab:rankings}
\begin{tabular}{|c|l|}
\hline
\textbf{Rank} & \textbf{COA1, COA2} \\ \hline
\multicolumn{2}{|c|}{\textbf{individual Dice coefficient (Eq. (3))}} \\ \hline
1 & 宣告分別財產制,聲請宣告夫妻分別財產制 \\ \hline
2 & 聲請宣告夫妻分別財產制,宣告夫妻分別財產制 \\ \hline
3 & 宣告改用分別財產制,聲請宣告夫妻分別財產制 \\ \hline
4 & 宣告分別財產制,宣告夫妻分別財產制 \\ \hline
\multicolumn{2}{|c|}{\textbf{wholistic Pearson’s correlation coefficient (Eq. (5))}} \\ \hline
1 & 宣告分別財產制, 宣告改用分別財產制 \\ \hline
2 & 宣告夫妻分別財產制, 宣告分別財產制 \\ \hline
3 & 宣告夫妻分別財產制, 宣告改用分別財產制 \\ \hline
4 & 聲請宣告夫妻分別財產制, 宣告分別財產制 \\ \hline
\multicolumn{2}{|c|}{\textbf{wholistic Dice coefficient (Eq. (6))}} \\ \hline
1 & 代位分割遺產, 代位請求分割遺產 \\ \hline
2 & 給付信用卡消費款, 給付簽帳卡消費款 \\ \hline
3 & 代位分割遺產, 代位請求分割共有物 \\ \hline
4 & 返還信用卡消費款, 清償現金卡消費款 \\ \hline
\end{tabular}
\end{table}

Here are the English translations for the Chinese terms in Table ~\ref{tab:rankings}. The translation for ``代位分割遺產", ``代位請求分割共有物", ``代位請求分割遺產", ``返還信用卡消費款", ``宣告分別財產制", ``宣告夫妻分別財產制", ``宣告改用分別財產制", ``清償現金卡消費款", ``給付信用卡消費款", ``給付簽帳卡消費款", and ``聲請宣告夫妻分別財產制" are, respectively, ``partition of inheritance by subrogation", ``subrogation claim for partition of common property", ``subrogation claim for partition of inheritance", ``refund of credit card expenses", ``declaration of separate property regime", ``declaration of separate property regime for spouses", ``declaration of change to separate property regime", ``settlement of cash card expenses", ``payment of credit card expenses", ``payment of debit card expenses", and ``application for declaration of separate property regime for spouses".

On the other hand, we may have anticipated that using different ranking scores would lead to different rankings among the similarity between COAs. Given 179 COAs, there are $179 \times 178 \div 2 = 15931$ different pairs of COAs, and it is not easy to ranked all of them with the same order. Our intuition is confirmed by the rankings in the three sub-tables in Table~\ref{tab:rankings} as well. Although the methods may have chosen the same pair in the top four, the actual rankings are different among them.

\subsection{Ensemble and Identiyfing Clusters of Similar COAs via Social Network Analysis}

We employed a simplistic ensemble model to combine the rankings of our models. We added the ranks reported by our three models, and preferred those had lower totalled ranks. We could then compute
the Pearson’s correlation coefficients (PCCs) between the rankings of the ensemble model and the rankings reported by the three base models. Table~\ref{tab:pcc} lists the observed PCCs. Overall, the PCCs look high.

\begin{table}[ht]
\centering
\caption{PCCs between the rankings of the base model and the ensemble model}
\label{tab:pcc}
\begin{tabular}{|l|c|}
\hline
\textbf{Model} & \textbf{PCC} \\ \hline
individual Dice coefficient (Eq. (3))   & 0.9603 \\ \hline
wholistic Pearson’s correlation coefficient (Eq. (5))   & 0.9092 \\ \hline
wholistic Dice coefficient (Eq. (6))    & 0.8791 \\ \hline
\end{tabular}
\end{table}

After this step, we could choose the top ranked pairs in the ensemble model, and adopted a tool for social network analysis (SNA), e.g., Gephi\footnote{Gephi: https://gephi.org/} to draw a social network. If some COAs were similar to each other, we expected that they would form clusters in the SNA graphs. Such clusters, if exist, would indicate more information about the similarities or relatedness among several COAs.

Figure 2 shows one of such clusters, when we used the top 100 pairs in the ensemble model to draw a social network. We converted the pairs that were recommended by our methods into node and link files that were needed by Gephi, and Gephi would find a social network of COAsfor us. In Figure 2, the numbered nodes represent “侵權行為損害賠償(交通)” (node 10), “損害賠償(交通)” (11), “侵權行為損害賠償(交通)” (19), “損害賠償(交通)” (24), “侵權行為損害賠償” (48), “請求侵權行為損害賠償” (49), and “損害賠償” (50). The COAs represented by these nine nodes form a common theme, although they did not for a clique. Nodes 10, 11, 19, and 24 form a clique, and nodes 20, 24, 48, 49, and 50 form another clique. These suggest that cases under these COAs are strongly related to each other in practice – supporting the similarity among the names of these COAs. 

\begin{figure}[h]
\includegraphics[width=1.5cm]{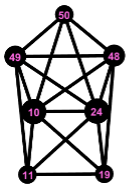}
\centering
\caption{A cluster drawn by Gehpi.}
\end{figure}

The cluster in Figure 2 was enlarged to become the red cluster in Figure 4, when we picked 1172 pairs to draw the social network. 
Readers may find its role in Figure 4 in the table in the Appendix.

\section{Determining the Similarity Threshold}

\subsection{Embedding and Clustering}

To analyze claims provided by plaintiffs, we embed and cluster these claims using the OpenAI-text-embedding-3-large model, which supports sequences up to 8192 tokens—often longer than the claims themselves. This model delivers embedding performance comparable to BERT-based tokenizers, such as the BERT-based Chinese model.

For clustering, we employ a density-based clustering algorithm, DBSCAN\footnote{DBSCAN in scikit learn: https://scikit-learn.org/stable/modules/generated/sklearn.cluster.DBSCAN.html}, as we do not have prior knowledge of the number of clusters. DBSCAN relies on two key parameters: epsilon ($\epsilon$) and minimum points ($minPts$). In the following, we describe our methodology for selecting these parameters.

Our goal is to ensure that cases with similar claims are closely situated in the embedded space. We acknowledge that while cases sharing the same COA may not always cluster together, their distribution across clusters holds significance. For instance, if COA1 is distributed across five clusters due to distinct prototypes, and COA2 is spread across the same five clusters, this suggests that COA2 shares certain characteristics with COA1, thereby indicating a higher probability of similarity between the two.

Liu \emph{et al.} define a ``more-similar-than" relationship using a related approach \cite{r6}. However, our work necessitates a more precise definition of ``exact similarity," requiring the assignment of a similarity score between cases. As described in Section 4, we use cited articles to rank similarity, and the ensemble model ranking introduced in Section 4.3 will inform the threshold determination for this task.

Inspired by the referenced work, we define similarity from a comparative perspective. Consider cases involving COA1 distributed across clusters $C_1$, $C_2$, $C_3$, and $C_4$, while COA2 is distributed across clusters $C_1$, $C_2$, $C_3$, and $C_5$. Both COAs appear in clusters $C_1$, $C_2$, and $C_3$. By counting the number of cases with each COA in these shared clusters and applying a predefined ratio standard, we can classify COAs as similar. Inheriting the notation from Section 2, let cases with COA $T_x$ belong to the set $\mathbf{S_x} = \{S_{x1}, S_{x2}, \ldots, S_{xi}, \ldots, S_{xn}\}$, and cases with COA $T_y$ belong to the set $\mathbf{S_y} = \{S_{y1}, S_{y2}, \ldots, S_{yj}, \ldots, S_{yn}\}$. After clustering, there are $s$ and $t$ cases in $\mathbf{S_x}$ and $\mathbf{S_y}$ within the same clusters, respectively. That is, for some $S_{xi}$, $S_{yj}$, and cluster $C_k$, $S_{xi} \in C_k$ and $S_{yj} \in C_k$. Given a specific standard $\lambda$, we define the similarity relation as follows in Eq. (7):

\begin{equation} 
\mathbf{R_{sim}} = \left\{ (T_x, T_y) \ \bigg| \ \frac{s}{n} \ge \lambda \text{ and } \frac{t}{n} \ge \lambda \right\},
\end{equation}
where $ 0 \le s \le n$ and $ 0 \le t \le n$. $\mathbf{R_{sim}}$ is the set of similar COA pairs selected by the DBSCAN process.

To ensure that all cases are assigned to appropriate clusters, we set $minPts = 1$. To determine the optimal combination of $\epsilon$ and $\lambda$, we define a utility function in the next subsection.

\subsection{Utility Function}

Our objective is to identify the optimal combination of $\epsilon$ and $\lambda$ such that the similarity relation $\mathbf{R_{sim}}$ includes as many similar COAs as possible. To achieve this, we aim to establish a threshold that effectively separates similar from non-similar COAs. As shown in Table~\ref{tab:pcc}, we observe that the individual Dice coefficient (Eq. (3)) has the highest Pearson correlation with the ensemble model ranking. Consequently, we incorporate the individual Dice coefficient into our utility function. Our primary approach is to compute the average individual Dice coefficient for all COAs within the similarity relation.

In addition to optimizing the accuracy of similarity identification, we aim to maximize the number of similar COAs identified by the utility function. To prevent the function from only selecting a few high-ranking COAs and neglecting others, we scale the average ranking by the total number of COAs in the similarity relation. This adjustment encourages the utility function to capture a broader range of COAs, including those with slightly lower rankings.

Let $\mathbf{T_{sim}}$ represent the set of COAs that are part of the similarity relation, and let $\mathbf{R_{sim}}$ denote the set of COA pairs that satisfy the similarity relation. The utility function is defined in Eq. (8) as follows, based on the epsilon ($\epsilon$) and standard ($\lambda$) parameters:

\begin{equation} U(\epsilon,\lambda) = |\mathbf{T_{sim}}|\frac{\sum_{(T_x,T_y) \in \mathbf{R_{sim}}}sim(T_x,T_y)}{|\mathbf{R_{sim}}|} \end{equation}

In this equation, $|\mathbf{T_{sim}}|$ represents the total number of COAs involved in the similarity relation, $sim((T_x,T_y))$ denotes the individual Dice coefficient introduced in Eq. (3) for each COA pair in $\mathbf{R_{sim}}$, and $|\mathbf{R_{sim}}|$ is the total number of COA pairs that satisfy the similarity relation. This utility function ensures a balance between the average individual Dice coefficient and the size of the COA set, ensuring that both the quality of the ranking and the number of relevant COAs are considered in the final utility score.

\subsection{Result and Threshold Determination}

Optimizing the utility defined in Eq. (8) is a process to find the best combination of $\lambda$ and $\epsilon$ for the DBSCAN step. 
The selection of $\epsilon$ affects the distribution of cases of a COA in the clusters that were produced by DBSCAN. The selection of $\lambda$ influences whether a pair of COAs will be considered similar based on the results of clustering, \emph{cf.} Eq. (7).

Since both s and t must be smaller than or equal to n, the sample size which we set to 200 in Sec. 4.2, the range of $\lambda$ is $[0, 1]$. Therefore, we may search for the best $\lambda$ from the list of $(0.005, 0.01, 0.015, \ldots, 1)$. DBSCAN uses $\epsilon$ as the upper bound of the distance between cases to put them in the same cluster, so this is usually a small number. We set minPts to 1 to permit isolated cases.

We employed a simulated annealing (SA)\footnote{Simulated Annealing in scipy:\\
https://docs.scipy.org/doc/scipy/reference/generated/scipy.optimize.dual\_annealing.html} procedure to find the best combination of $\lambda$ and $\epsilon$, letting the SA procedure aim to optimize the utility, defined in Eq. (8).

After completing the optimization process, we found that the combination $\epsilon = 0.4376$ and $\lambda = 0.01$ yields the best utility score of 16.83. Figure 3 illustrates the distribution of our results in determining the similarity threshold. In total, our process identified 47 similar COA pairs. However, the three COA pairs with the lowest rankings—ranked 2180, 4291, and 8826—are notably separated from the other 44 pairs. We recognize these three as outliers and exclude them from the threshold selection process.

\begin{figure}[h] 
\includegraphics[width=8cm]{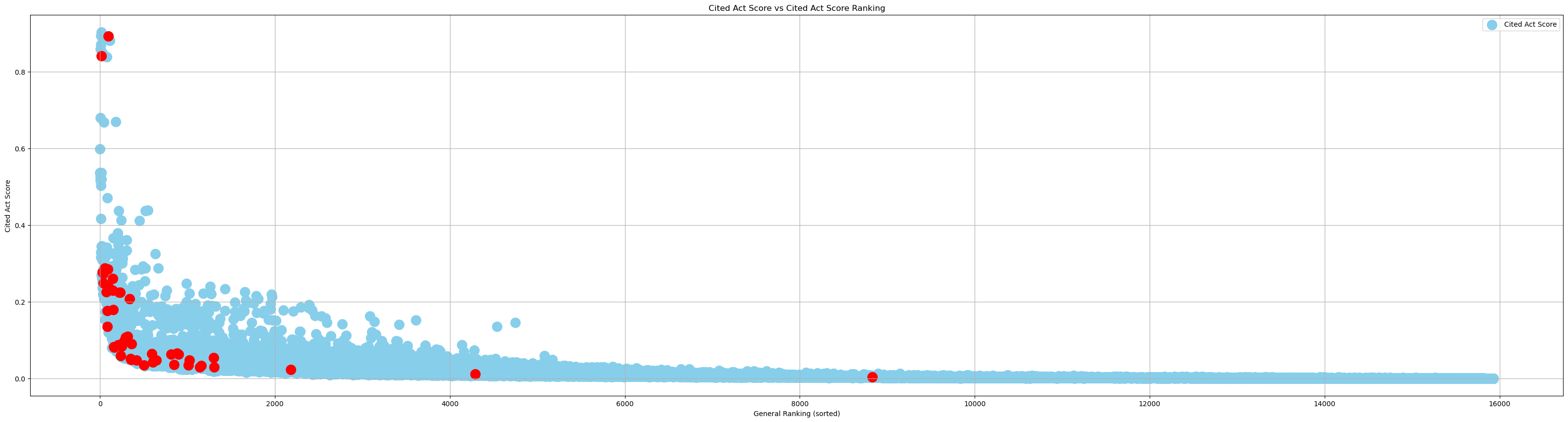} 
\centering 
\caption{Distribution of similar COA pairs and their ensemble model rankings} 
\end{figure}

For the remaining 44 COA pairs, the average ensemble ranking is $\mu = 412$, with a standard deviation of $\sigma = 380$. To establish a statistically supported threshold, we calculate $\mu + 2\sigma$. Thus, we set our threshold for similarity at 1172, recognizing any COA pairs ranked above this value as similar.

\section{Empirical Observations}

Using the social network analysis method described in Section 4.3, we employed Gephi to visualize the COA pairs ranked above 1172, as shown in Figure 4. In this graph, each number on a node represents a specific COA. The size of the node corresponds to its connectivity, while the color reflects the results of statistical inferences generated by Gephi.

\begin{figure}[h]
\includegraphics[width=12cm]{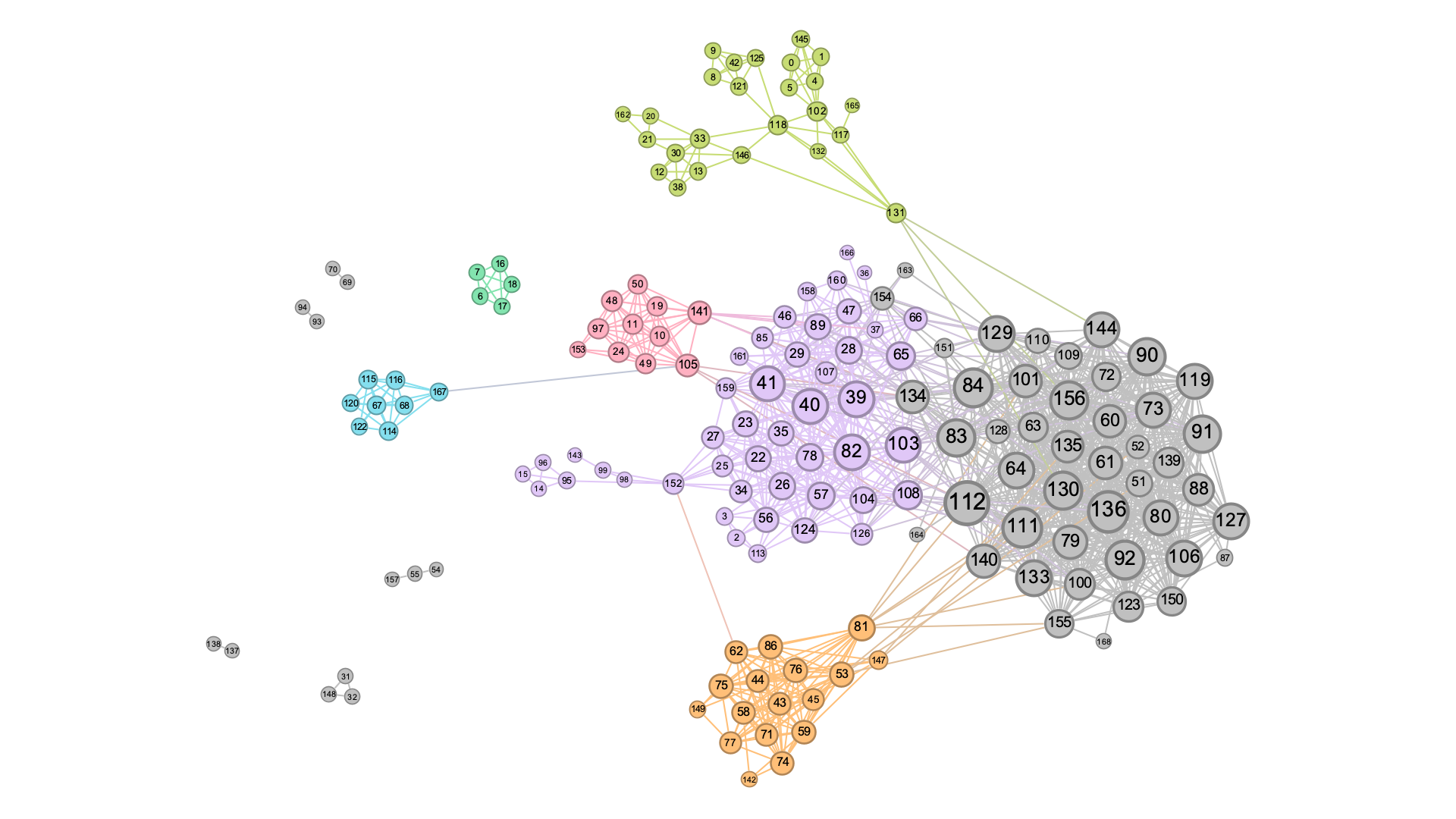} 
\centering
\caption{Social network representation of the similarity relationships between COAs} 
\end{figure}

Upon examining the graph, we observe that each cluster carries distinct legal meanings. For instance, the cluster containing COA numbers 6, 7, 16, 17, and 18 all pertain to marital property. Similarly, the cluster consisting of COA numbers 67, 68, 114, 115, 116, 120, 122, and 167 is related to labor disputes, addressing issues such as pensions (67, 68), salaries (114, 115, 116), overtime pay (120), compensation for occupational accidents (167), and the existence of employment relationships (122).

Furthermore, certain COAs are connected to multiple clusters, such as COA numbers 83, 84, and 134. This overlap can also be explained: COAs 83 and 84 relate to unjust enrichment, which is linked both to matters of repayment (the large purple cluster) and matters of immovable properties (the large gray cluster), while COA 134 pertains to status restoration, which ties to both clusters.

A detailed explanation of each cluster and its corresponding legal meaning is provided in the Appendix.

\section{Discussion and Concluding Remarks}

We have reported the analysis based on the setting of (m, n) = (179, 200), \emph{cf.} Sections 2 and 4.2. We have conducted another run of analysis for (m, n)=(139,300). Very similar results were achieved, where the similarity between the rankings observed in different settings was measured by Pearson's correlation coefficient.  

The proposed method shows that even without providing the machine with predefined COA names, it can still cluster similar COAs based on the Dice coefficient of cited articles and basic fact embeddings. This expands legal analysis by uncovering connections between COAs that may not be immediately apparent through traditional methods, addressing the challenges about the different aspects of defining similarity between cases that were discussed in a JURIX 2023 workshop \cite{r7}.

Defining similarity and relevance between cases is certainly not straightforward \cite{r8}. They may be defined computationally in different ways for different needs in a certain circumstance or under different computational scenarios, e.g., \cite{r9}. Similarity can be defined based on court decisions \cite{r10} or based on the nature of disputes in civil cases \cite{r6}.

The approach reported in this presentation could be applied broadly across various branches of Taiwanese law, such as criminal and administrative law. There is no obvious reason that prevents its application to other languages. By utilizing machine learning and unsupervised clustering, it offers more comprehensive reachability to legal research, overcoming the limits of manually assigned legal categories.


\section*{Appendix: Clusters and Corresponding Legal Meanings}
\begin{table}[h!]
\centering
\begin{tabular}{|c|l|}
\hline
\textbf{Nodes} & \textbf{Description} \\ \hline
[0,1,4,5,102,117,118,132,145,165] & Existence and partition of a succession \\ \hline
[8,9,42,121,125] & Existence of a parent-child relation \\ \hline
[12,13,30,33,38,146] & Existence of a marriage relation \\ \hline
[20,21,162] & Divorce \\ \hline
[6,7,16,17,18] & Marital properties \\ \hline
[69,70] & Right of way by necessity \\ \hline
[93,94] & Boundaries of immovable property \\ \hline
[137,138] & State Compensation \\ \hline
[31,32,148] & Existence of mandate relationships of directors \\ \hline
[54,55,157] & Related to attachments \\ \hline
[98,99,143] & Debtor objection suit \\ \hline
[14,15,95,96] & Related to negotiable instruments \\ \hline
Cyan color & Related to labor disputes \\ \hline
Red color & Torts \\ \hline
Yellow color & Loan repayment \\ \hline
Other purple color & Other repayment \\ \hline
Other gray color & Related to immovable properties \\ \hline
\end{tabular}
\end{table}

\end{CJK}

\end{document}